%% file: main.tex
\documentclass[10pt,twocolumn,letterpaper]{article}

\usepackage{iccv}
\usepackage{times}
\usepackage{epsfig}
\usepackage{graphicx}
\usepackage{amsmath}
\usepackage{amssymb}
\usepackage{multirow}

\input{macros}

\usepackage[pagebackref=true,breaklinks=true,letterpaper=true,colorlinks,bookmarks=false]{hyperref}

\iccvfinalcopy 

\def\iccvPaperID{2429} 

\ificcvfinal\fi
\begin{document}

\title{Extreme clicking for efficient object annotation}

\author{\hspace{-0.8cm} Dim P. Papadopoulos\textsuperscript{1} \hspace{1.2cm} Jasper R. R. Uijlings\textsuperscript{2} \hspace{1.0cm} Frank Keller\textsuperscript{1} \hspace{1.7cm} Vittorio Ferrari\textsuperscript{1,2} \\
{\tt\small \hspace{-0.8cm} dim.papadopoulos@ed.ac.uk \hspace{0.4cm} jrru@google.com \hspace{0.15cm} keller@inf.ed.ac.uk \hspace{0.2cm} vferrari@inf.ed.ac.uk}\\
\textsuperscript{1}University of Edinburgh  \hspace{1cm} \textsuperscript{2}Google Research
}

\maketitle


\input{secAbstract}
\input{secIntro}
\input{secRelWork}

\input{secMturk}

\input{secSegMethod}
\input{secResultsTradeoff}
\input{secResultsSeg}

\input{secConclusions}

\input{secAppendix}

{\small
\bibliographystyle{ieee}
\bibliography{../../../bibtex/shortstrings,../../../bibtex/vggroup,../../../bibtex/calvin,../../../bibtex/viscog}
}

\end{document}

%% file: macros.tex
\iftrue     
\newcommand{\dimpp}[1]{\textcolor{blue}{[DP: #1]}}
\newcommand{\jasper}[1]{\textcolor{magenta}{[JU: #1]}}
\newcommand{\vitto}[1]{\textcolor{red}{[VF: #1]}}
\newcommand{\frank}[1]{\textcolor{cyan}{[FK: #1]}}
\newcommand{\arxiv}[1]{\textcolor{green}{[Arxiv: #1]}}
\else
\newcommand{\dimpp}[1]{\textcolor{blue}{\noindent}}
\newcommand{\jasper}[1]{\textcolor{magenta}{\noindent}}
\newcommand{\vitto}[1]{\textcolor{red}{\noindent}}
\newcommand{\frank}[1]{\textcolor{cyan}{\noindent}}
\newcommand{\arxiv}[1]{\textcolor{green}{\noindent}}
\fi

\newcommand{\mypar}[1]{\vspace{-5mm}\paragraph{#1}}

\DeclareMathAlphabet{\mathpzc}{T1}{pzc}{m}{n}

\newlength{\halfwidth}
\newlength{\fullwidth}
\newlength{\tikzimgheight}
\newlength{\tikzimgwidth}

\usepackage{ifthen}

%% file: secAbstract.tex
\begin{abstract}

Manually annotating object bounding boxes is central to building computer vision datasets,
and it is very time consuming (annotating ILSVRC~\cite{russakovsky15ijcv} took 35s for one high-quality box~\cite{su12aaai}).
It involves clicking on imaginary corners of a tight box around the object. This is difficult as these corners are often outside the actual object and several adjustments are required to obtain a tight box.
We propose extreme clicking instead: we ask the annotator to click on four physical points on the
object: the top, bottom, left- and right-most points. This task is more natural and these points are easy to find.
%
We crowd-source extreme point annotations for PASCAL VOC 2007 and 2012 and show that
(1) annotation time is only 7s per box, $5\times$ 
faster than the traditional way of drawing boxes~\cite{su12aaai};
(2) the quality of the boxes is as good as the original ground-truth drawn the traditional way; 
(3) detectors trained on our annotations are as accurate as those trained on the original ground-truth.
Moreover, our extreme clicking strategy not only yields box coordinates, but also four accurate boundary points.
We show
(4) how to incorporate them into GrabCut to obtain more accurate segmentations than those delivered when initializing it from bounding boxes;
(5) semantic segmentations models trained on these segmentations outperform those trained on segmentations derived from bounding boxes.

\end{abstract}

%% file: secIntro.tex
\section{Introduction}
\label{sec:intro}

\begin{figure}[t]
\vspace{-.1cm}
\center
\includegraphics[width=\linewidth]{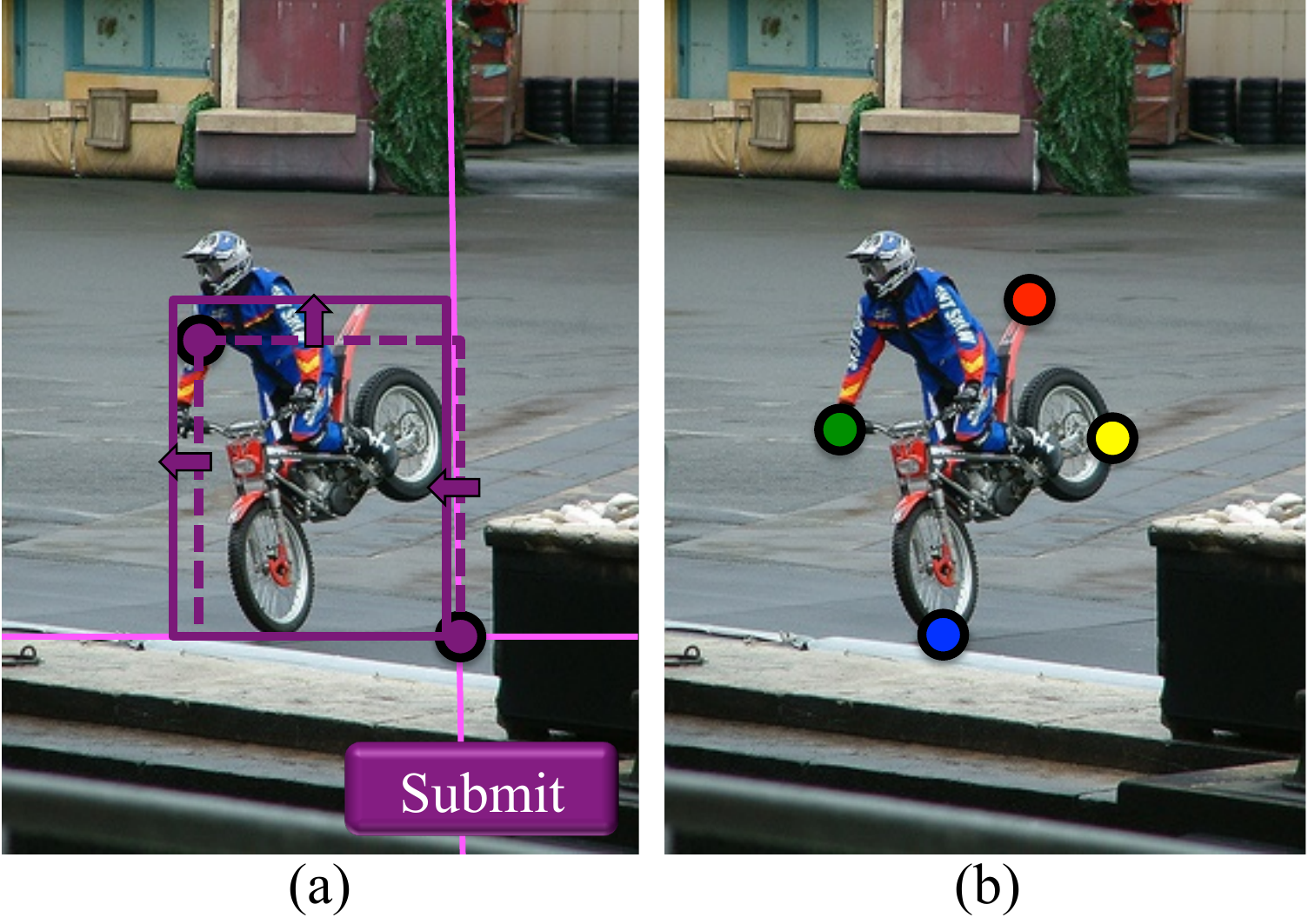}
\vspace{-.4cm}
\caption{\small \textbf{Annotating an instance of motorbike:} (a) The conventional way of drawing a bounding box. (b) Our proposed extreme clicking scheme.} 
\vspace{-.5cm}
\label{fig:splash}
\end{figure}

Drawing the bounding boxes traditionally used for object detection
is very expensive. The PASCAL VOC bounding boxes were obtained by
organizing an ``annotation party'' where expert annotators 
 were gathered in one place to create high
 quality annotations~\cite{everingham10ijcv}.
%
But crowdsourcing is essential for creating larger datasets:
Su et al.~\cite{su12aaai} developed an efficient protocol to annotate high-quality boxes using
Amazon Mechanical Turk (AMT). They
report 39\% efficiency gains over consensus-based approaches (which collect multiple
annotations to ensure quality)~\cite{deng09cvpr,sorokin08cvprw}. However, even this
efficient protocol requires 35s to annotate one box (more details in Sec.~\ref{sec:relwork}).

Why does it take so long to draw a bounding box?
Fig~\ref{fig:splash}a shows the typical
process~\cite{crowdflowerboxannotation16,everingham10ijcv,jain13iccv,russakovsky15cvpr,spare5boxannotation17,su12aaai}.
First the annotator clicks on a corner of an imaginary
rectangle tightly enclosing the object (say the bottom-right corner). 
This is challenging, as these corners are typically not on the object.
Hence the annotator needs to find the relevant extreme points
of the object (the bottom point and the rightmost point) and
adjust the x- and y-coordinates of the corner to match them.
After this, the annotator clicks and drags the mouse to the diagonally opposite corner.
This involves the same process of x- and y-adjustment, but now based on a visible
rectangle. After the rectangle is adjusted, the annotator clicks again. 
He/she can make further adjustments by clicking on the sides of the rectangle and dragging them
until the box is tight on the object.
Finally, the annotator clicks a ``submit'' button.

From a cognitive perspective, the above process is
suboptimal. The three steps (clicking on the first corner,
dragging to the second corner, adjusting the sides) effectively
constitute three distinct tasks. Each task requires attention to
different parts of the object and using
the mouse differently. In effect, the annotator is constantly
task-switching, a process that is cognitively demanding
and is correlated with increased response times
and errors rates \cite{Monsell:03,Rubinstein:ea:01}.
Furthermore, the process involves a substantial amount
of mental imagery: the rectangle to be drawn is imaginary, and so are the
corner points. 
Mental imagery also has a cognitive cost, e.g. in mental rotation
experiments, response time is proportional to rotation
angle \cite{Kosslyn:ea:95,Shepard:Metzler:71}.

In this paper we propose an annotation scheme which
avoids task switching and mental imagery, 
resulting in greatly improved efficiency. 
We call our scheme \emph{extreme clicking}: we ask the annotator to click on four
extreme points of the object, i.e. points belonging to the
top, bottom, left-most, and right-most parts of the object (Fig~\ref{fig:splash}b).
This has several advantages:
(1) Extreme points are not imaginary, but are
well-defined physical points on the object, which makes them easy to locate.
(2) No rectangle is
involved, neither real nor imaginary. This further reduces mental
imagery, and avoids the need for detailed instructions defining the
notion of a bounding box.
(3) Only a single task is
performed by the annotator
thus avoiding task switching.
(4) No separate box adjustment step is required.
(5) No ``submit'' button is necessary; annotation terminates after four clicks.

Additionally, extreme clicking provides \emph{more information} than just box coordinates:
we get four points on the actual object boundary.
We demonstrate how to incorporate them into GrabCut~\cite{Rother04vitto}, to deliver more accurate segmentations than
when initializing it from bounding boxes~\cite{Rother04vitto}.
In particular, GrabCut relies heavily on the initialization of the object appearance model (e.g.~\cite{kuettel12cvpr,Rother04vitto,WangICCV05}) and on which
pixels are clamped to be object/backgound.
When using just a bounding box, the object appearance model is initialized from all pixels within the box
(e.g.~\cite{ferrari08cvpr,kuettel12cvpr,Rother04vitto}).
Moreover, it typically helps to clamp a smaller central region to be object~\cite{ferrari08cvpr}.
Instead, we first expand our four object boundary points to an estimate of the whole contour of the object.
We use this estimate to initialize the GrabCut object appearance model. Furthermore, we skeletonize the estimate and clamp the
resulting pixels to be object.

We perform extensive
experiments on PASCAL VOC 2007 and 2012 
using crowd-sourced annotations which demonstrate:
(1) extreme clicking only takes 7s seconds per box, $5\times$ faster than the traditional way of drawing boxes~\cite{su12aaai};
(2) extreme clicking leads to high-quality boxes on a par with the original ground-truth boxes drawn the traditional way;
(3) detectors trained on boxes generated using extreme clicking perform as well as those trained on the original ground-truth;
(4) incorporating extreme points into GrabCut \cite{Rother04vitto} improve object segmentations by 2\%-4\% mIoU over initializing it from bounding boxes;
(5) semantic segmentations models trained on segmentations derived from extreme clicking outperform those trained on segmentations generated from bounding boxes by 2.6\% mIoU.

%% file: secRelWork.tex
\section{Related work}
\label{sec:relwork}

\begin{figure*}[t]
\center
\vspace{-.4cm}
\includegraphics[width=\linewidth]{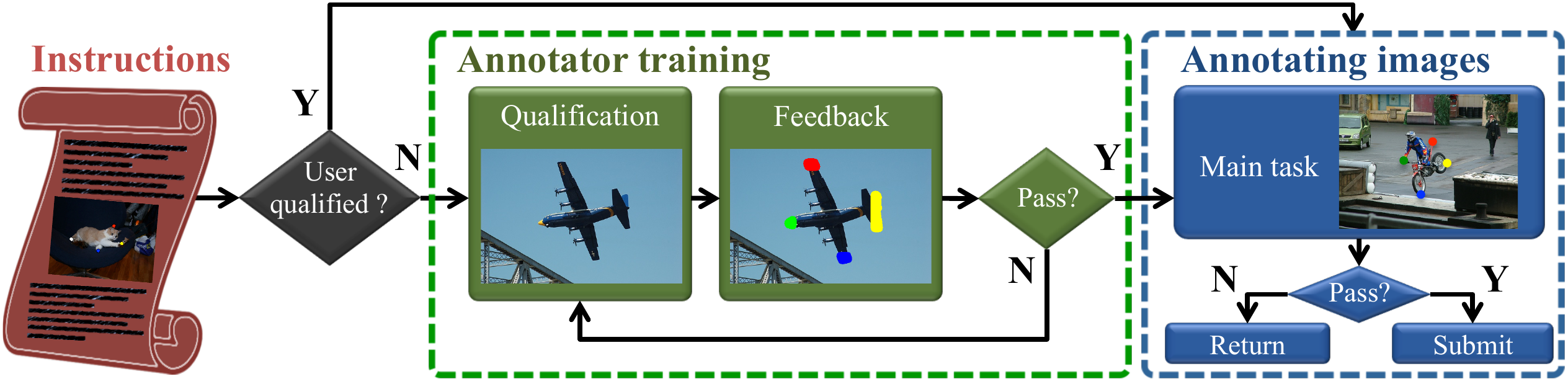}
\vspace{-.4cm}
\caption{\small \textbf{The workflow of our crowd-sourcing protocol for collecting extreme click annotations on images.} The annotators read a set of instructions and then go through an interactive training stage that consists of a qualification test at the end of which they receive a detailed feedback on how well they performed. Annotators who successfully pass the test can proceed to the annotation stage. In case of failure, they are allowed to repeat the test as many times as they want until they succeed.}
\vspace{-.4cm}
\label{fig:mturkflow}
\end{figure*}

\vspace{-.1cm}
\paragraph{Time to draw a bounding box.}
The time required to draw a bounding box varies depending on several factors, including the quality of the boxes and the crowdsourcing protocol used.
In this paper, as an authoritative reference we use the protocol of~\cite{su12aaai} which was used to annotate ILSVRC~\cite{russakovsky15ijcv}.
It was designed to produce high-quality bounding boxes with little
human annotation time on Amazon Mechanical Turk.
They report the following median times for annotating an object of a given class in an image~\cite{su12aaai}:
25.5s for drawing one box,
9.0s for verifying its quality,
and 7.8s for checking whether there are other objects of the same class yet to be annotated.
Since we only consider annotating one object per class per image, we use $25.5s + 9.0s = 34.5s$ as the reference time.
This is a conservative estimate:
when taking into account that some boxes are rejected and need to be re-drawn, the median time increases to 55s. If we use average times instead of medians, the cost raises further to 117s.

Note how both PASCAL VOC and ILSVRC have images of comparable difficulty and come with ground-truth box annotations of similar high quality~\cite{russakovsky15ijcv}, justifying our choice of 35s reference time.
Papers reporting faster timings~\cite{jain13iccv,russakovsky15cvpr} aim for lower-quality boxes (e.g. the official annotator instructions of~\cite{jain13iccv} show an example box which is not tight around the object).
We compare to~\cite{russakovsky15cvpr} in Sec.~\ref{sec:resultsDet}.

\mypar{Reducing annotation time for training object detectors.}
Weakly-supervised object localization techniques (WSOL) can be used to train object detectors from image-level labels only (without bounding boxes)~\cite{bilen16cvpr,cinbis15pami,deselaers12ijcv,russakovsky12eccv,siva11iccv}. This setting is very cheap in terms of annotation time, but it produces lower quality object detectors, typically performing at only about half the level of accuracy achieved by training from bounding boxes~\cite{bilen16cvpr,cinbis15pami,deselaers12ijcv,russakovsky12eccv,wang15tip}.

Training object class detectors from videos could bypass the need for manual bounding boxes, as the motion of the objects facilitates their automatic localization~\cite{prest12cvpr,singh16cvpr,kuznetsova15cvpr}. However, because of the domain adaptation problem, these detectors are still quite weak compared to ones trained on manually annotated still images~\cite{kalogeiton16pami}.
Alternative types of supervision information such as eye-tracking data~\cite{mathe13nips,papadopoulos14eccv}, text from news articles or web pages~\cite{DuyguluECCV02,Gupta08:eccv}, or even movie scripts~\cite{bojanowski13iccv} have also been explored.
%
Papadopoulos et al.~\cite{papadopoulos16cvpr} propose a scheme for training object class detectors which only requires annotators to verify bounding boxes generated automatically by the learning algorithm.
We compare our extreme clicking scheme to state-of-the-art WSOL~\cite{bilen16cvpr}, and to~\cite{papadopoulos16cvpr} in Sec.~\ref{sec:resultsDet}.

\mypar{(Interactive) object segmentation.}
Object segmentations are significantly more expensive to obtain than
bounding boxes. The creators of the SBD
dataset~\cite{hariharan11iccv} merged five annotations per instance, resulting in a total time of 315s per instance. For COCO~\cite{lin14eccv}, 79s per instance were required for
drawing object polygons, excluding verifying correctness and possibly redrawing.
%
To reduce annotation time many interactive segmentation techniques have been proposed, which require the user to input either a bounding box around the object
\cite{lempitsky:iccv09,Rother04vitto,wu14cvpr}, or scribbles \cite{bai09ijcv,duchenne08cvpr,freedman05cvpr,GradyPAMI06,gulshan10cvpr,lempitsky:iccv09,price10cvpr,veksler08eccv,VicenteCVPR08,yang10tip},
or clicks~\cite{jain16hcomp,wang14cviu}.
Most of this work is based on the seminal GrabCut algorithm~\cite{Rother04vitto}, which iteratively
alternates between estimating appearance models (typically Gaussian Mixture
Models~\cite{BlakeECCV04}) and refining the segmentation using graph cuts~\cite{boykov:pami04}.
The user input is typically used to initialize the appearance model and to clamp some pixels to background.
%
In this paper, we incorporate extreme clicks into GrabCut~\cite{Rother04vitto}, improving the appearance model initialization
and automatically selecting good seed pixels to clamp to object.




%% file: secMturk.tex
\section{Collecting extreme clicks}
\label{sec:mturk}

In this section, we describe our crowd-sourcing framework for collecting extreme click annotations (Fig.~\ref{fig:mturkflow}).
Annotators read a simple set of instructions (sec.~\ref{sec:mturk_instr}) and then go through an interactive training stage (sec.~\ref{sec:mturk_train}). Those who successfully pass the training stage can proceed to the annotation stage (sec.~\ref{sec:mturk_anno}). 

\subsection{Instructions}
\label{sec:mturk_instr}

The annotators are given an image and the name of a target object class.
They are instructed to click on four extreme points (top, bottom, left-most, right-most) on the visible part 
of any object of this class.
They can click the points in any order.
In order to let annotators know approximately how long the task will take, we suggest a total time of 10s for all four clicks. This is an upper bound on the expected annotation time that we estimated from a small pilot study.

Note that our instructions are extremely simple, much simpler than those necessary
to explain how to draw a bounding box in the traditional way
(e.g.~\cite{russakovsky15cvpr,su12aaai}).
%
%
They are
also simpler than instructions required for verifying whether a displayed bounding box is
correct~\cite{papadopoulos16cvpr,russakovsky15cvpr,su12aaai}. That requires the annotator to imagine
a perfect box on the object, and to mentally compare it to the displayed one.

\subsection{Annotator training}
\label{sec:mturk_train}

After reading the instructions, the annotators go through the training stage. They have to complete a qualification test, at the end of which they receive detailed feedback on how well they performed. Annotators who successfully pass this test can proceed to the annotation stage. In case of failure, annotators can repeat the test until they succeed.

\mypar{Qualification test.}
A qualification test is a good mechanism for enhancing the quality of crowd-sourcing data and for filtering out bad annotators and spammers~\cite{andriluka14cvpr,endres10cvprw,Johnson11cvpr,krause2013iccvw}. Some annotators do not pay attention to the instructions or do not even read them.
Qualification tests have been successfully used to collect image labels, object bounding boxes, and segmentations for some of the most popular datasets (e.g.,~COCO~\cite{lin14eccv} and Imagenet~\cite{russakovsky15ijcv,su12aaai}).

The qualification test is designed to mimic our main task of clicking on the extreme points of objects.
We show the annotator a sequence of $5$ different images with the same object class and ask
them to carry out the extreme clicking task.

\mypar{Feedback.}
The qualification test uses a small pool of images with ground-truth segmentation masks for the
objects, which we employ to automatically validate the annotator's clicks and to provide feedback
(Fig.~\ref{fig:mturkflow}, middle part). We take a small set of qualification images from a different dataset than
the one that we annotate.

\begin{figure}[t]
\vspace{-.1cm}
\center
\includegraphics[width=\linewidth]{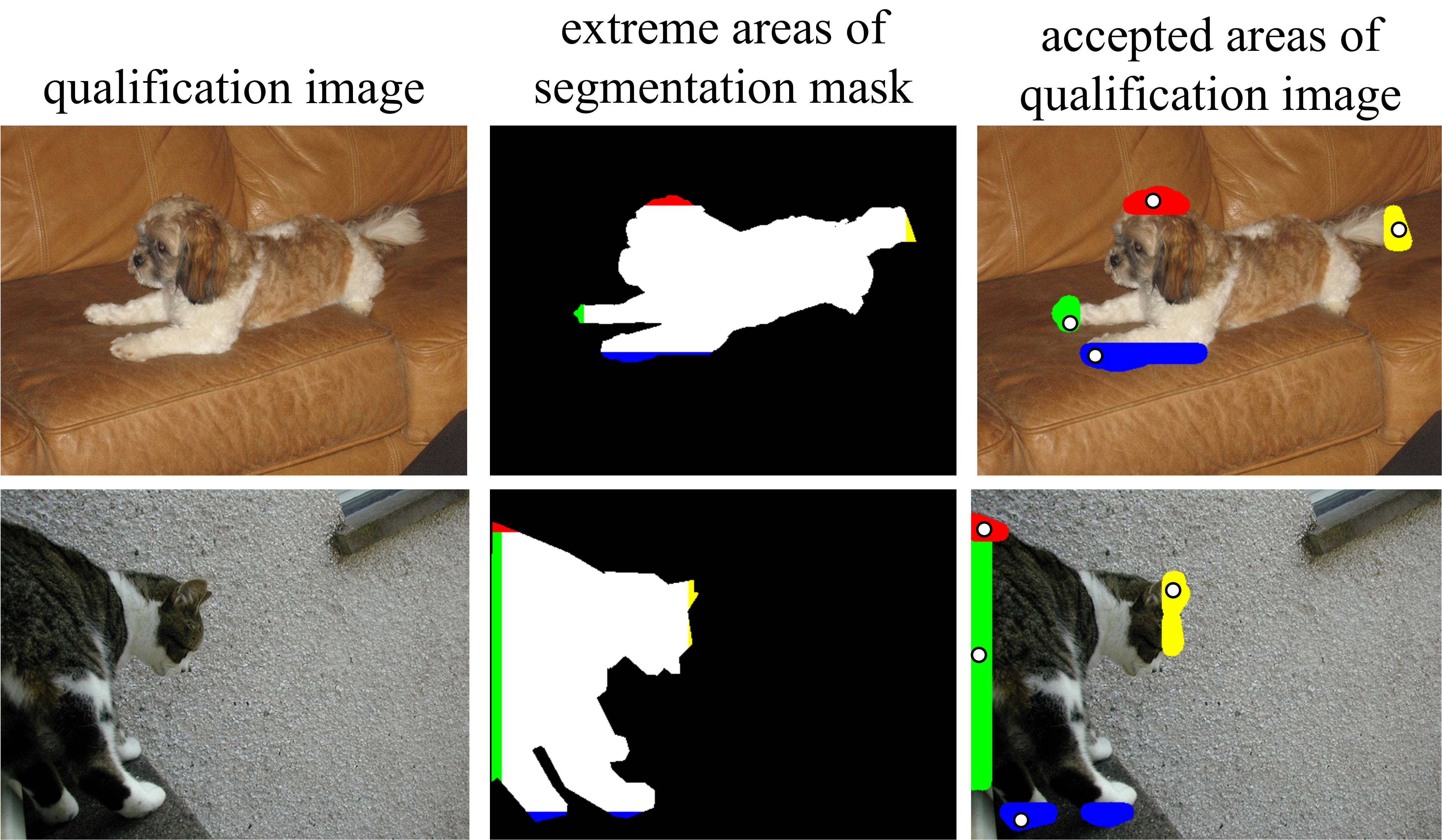}
\vspace{-.4cm}
\caption{\small \textbf{Qualification test.} (Left) Qualification test examples of the dog and cat class. (Middle) The figure-ground segmentation masks we use to evaluate annotators' extreme clicks during the training stage. The pixels of the four extreme areas of the mask are marked with colors. (Right) The accepted areas for each extreme click and the click positions as we display them to the annotators as feedback.}
\vspace{-.5cm}
\label{fig:qualTest}
\end{figure}

In the following, we explain the validation procedure for the top click (the other three cases are analogous).
We ask the annotator to click on a top point on the object, but this point is not necessarily uniquely defined. Depending on the object shape, there may be multiple points that are equivalent, up to some tolerance margin (e.g. the top of the dog's head in fig.~\ref{fig:qualTest}, top row). Clearly, clicking on any of these points is correct.
The area in which we accept the annotator's click is derived from the segmentation mask. 
First, we find the pixels with the highest y-coordinate in it (there might be multiple such pixels).
Then, we select all pixels in the mask with y-coordinates within $10$ pixels of any of these top pixels (red area in Fig.~\ref{fig:qualTest}, middle column).
Finally, we also include in the accepted area all image pixels within $10$ pixels of any of the selected pixels in the segmentation mask (Fig.~\ref{fig:qualTest}, right column).
Thus the accepted area includes all top pixels in the mask, plus a tolerance region around them, both inside and outside the mask.

After the annotators finish the qualification test, they receive a feedback page with all the examples they annotated. For each image, we display the annotator's four clicks, and the accepted areas for each click (Fig.~\ref{fig:qualTest} right column).

\mypar{Success or failure.}
The annotators pass the qualification test if all their clicks on all 5 qualification images are inside the accepted areas. Those that pass the test are recorded as qualified annotators and can proceed to the main annotation stage. A qualified annotator never has to retake the qualification test. In case of failure, annotators are allowed to repeat the test as many times as they want. The combination of automatically providing rich feedback and allowing annotators to repeat the test makes the training stage interactive and highly effective. Annotators that have reached the desired level of quality can be expected to keep it throughout the annotation~\cite{hata17cscw}.

\subsection{Annotating images}
\label{sec:mturk_anno}

In the annotation stage, annotators are asked to annotate small batches of $10$ consecutive images.
To increase annotation efficiency, the target class for all the images within a batch is the same. This means annotators do not have to re-read the class name for every image and can use their prior knowledge of the class to find it rapidly in the image~\cite{Torralba:ea:06}. More generally, it avoids task-switching which is well-known to increase response time and decrease accuracy~\cite{Rubinstein:ea:01,Monsell:03}.

\mypar{Quality control.}
Quality control is a common process when crowd-sourcing image annotations~\cite{bearman16eccv,kovashka15ijcv,lin14eccv,russakovsky15ijcv,russell08ijcv,sorokin08cvprw,su12aaai,vondrick13ijcv,welinder10nips}.
We control the quality of the annotation by hiding one evaluation image for which we have a ground-truth segmentation inside a 10-image batch, and monitor the annotator's accuracy on it (golden question). Annotators that fail to click inside the accepted areas on this evaluation image are not able to submit the task. 
We do not do any post-processing rejection of the submitting data.

%% file: secSegMethod.tex
\section{Object segmentation from extreme clicks}
\label{sec:segMethod}

\begin{figure*}[t]
\center
\vspace{-.4cm}
\includegraphics[width=\linewidth]{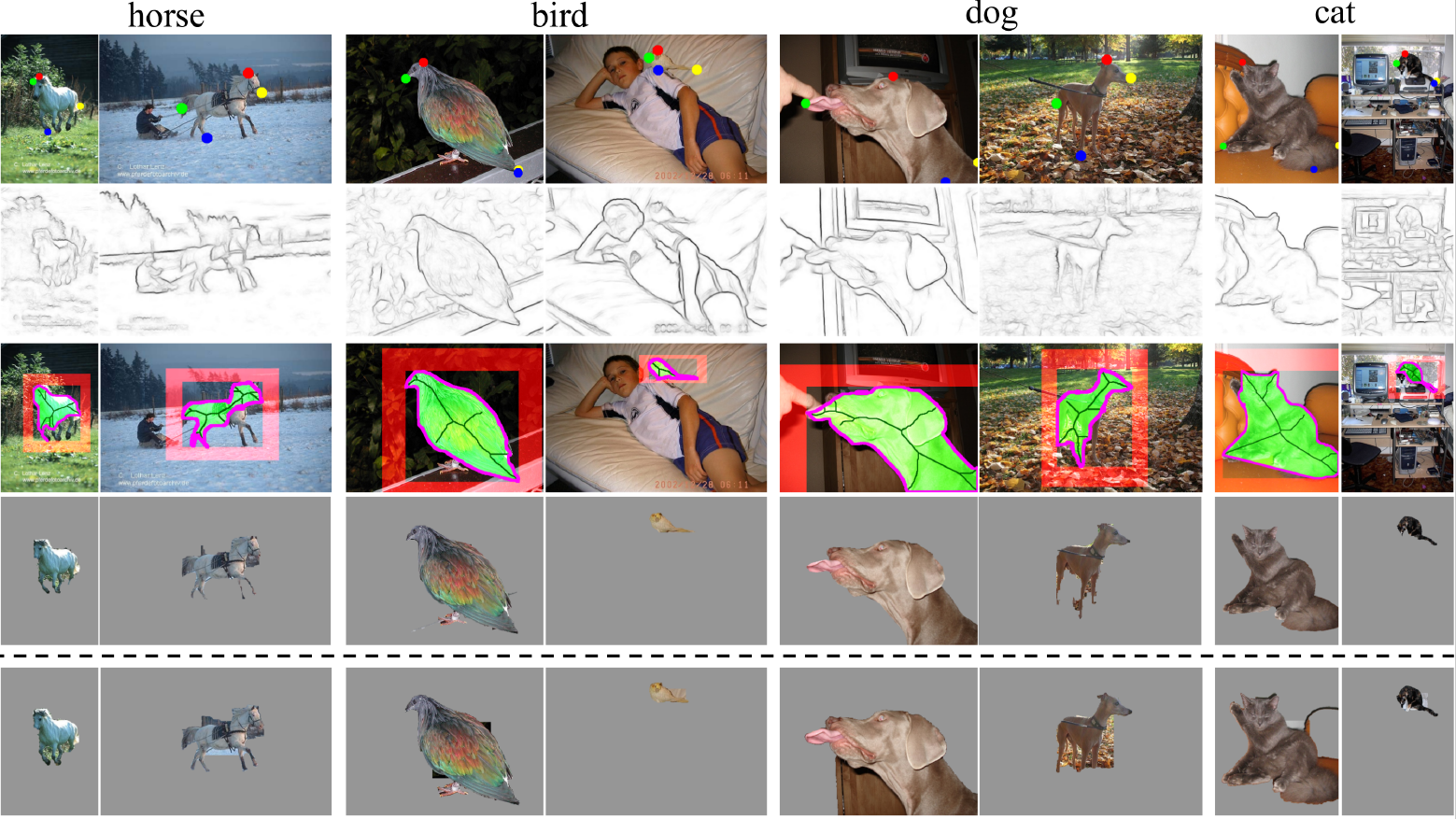}
\vspace{-.4cm}
\caption{\small \textbf{Visualization of input cues and output of GrabCut.} First row shows input with annotator's extreme clicks. Second row shows output of edge detector~\cite{dollar13iccv}. Third row shows our inputs for GrabCut:  the pixels used to create background appearance model (red), the pixels used to create the object appearance model (bright green), the initial boundary estimate (magenta), and the skeleton pixels which we clamp to have the object label (dark green). Fourth row shows the output of GrabCut when using our new inputs, while the last row shows the output when using only a bounding box.}
\vspace{-.4cm}
\label{fig:boundApp}
\end{figure*}

Extreme clicking results not only in high-quality bounding box annotations, but also in four accurate object boundary points. In this section we explain how we use these boundary points to improve the creation of segmentation masks from bounding boxes.

We cast the problem of segmenting an object instance in image $I$ as a pixel labeling problem. Each pixel $p \in I$ should be labeled as either object ($l_{p}=1$) or background ($l_{p}=0$). A labeling $L$ of all pixels represents the segmented object. 
Similar to~\cite{Rother04vitto}, we employ a binary pairwise energy function $E$ defined over the pixels and their labels.
\begin{equation}
\label{eq:energyGC}
E(L) = \sum_p{U(l_p) } + \sum_{p,q}{V (l_p,l_q) }
\end{equation}
$U$ is a unary potential that evaluates how likely a pixel $p$ is to take label $l_p$ according to the object and background appearance models, while the pairwise potential $V$ encourages smoothness by penalizing neighboring pixels taking different labels.


\mypar{Initial object surface estimate from extreme clicks.}
For GrabCut to work well, it is important to have a good initial estimate of the object surface to initialize the appearance model. Additionally, it helps to clamp certain pixels to object~\cite{kuettel12cvpr}. We show how the four collected object boundary points can be exploited to do both. 

In particular, for each pair of consecutive extreme clicks (e.g. leftmost-to-top, or top-to-rightmost) we find the path connecting them which is most likely to belong to the object boundary. For this purpose we first apply a strong edge detector~\cite{dollar13iccv}
to obtain a boundary probability $e_p \in [0,1]$ for every pixel $p$ of the image (second row of Fig.~\ref{fig:boundApp}).
We then define the best boundary path between two consecutive extreme clicks as the shortest path whose minimum edge-response is the highest (third row of Fig.~\ref{fig:boundApp}, magenta). 
We found this objective function to work better than others, such as minimizing $\sum_p (1-e_p)$ for pixels $p$ on the path.
The resulting object boundary paths yield an initial estimate of the object outlines.

We use the surface within the boundary estimates (shown in green in the third row of Fig.~\ref{fig:boundApp}) to initialize the object appearance model used for $U$ in Eq.~\eqref{eq:energyGC}. Furthermore, from this surface we obtain a skeleton using standard morphology (shown in dark green in third row of Fig.~\ref{fig:boundApp}). This skeleton is very likely to be object, so we clamp its pixel-labels to be object ($l_s = 1$ for all pixels $s$ on the skeleton).

\mypar{Appearance model.}
As in classic GrabCut~\cite{Rother04vitto}, the appearance model consists of two GMMs, one for the object (used when $l_{p}=1$) and one for the background (used when $l_{p}=0$). Each GMM has five components, where each is a full-covariance Gaussian over the RGB color space.

Traditional interactive segmentation techniques~\cite{lempitsky:iccv09,Rother04vitto,wu14cvpr} start from a manually drawn bounding box and estimate the initial appearance models from all pixels inside the box (object model) and all pixels outside it (background model). However, this may be suboptimal: since we are trying to segment the object within the box, intuitively only the immediate background is relevant, not the whole image. Indeed, we improved results by using a small ring around the bounding box for initializing the background model (see third row Fig.~\ref{fig:boundApp} in red).
Furthermore, not all pixels within the box belong to the object. But given only a bounding box as input, the best is to still use
the whole box to initialize the object model.
Therefore, in our baseline GrabCut implementation, the background model is initialized from the immediate background and the object model is initialized from all pixels within the box.

\begin{figure}[t]
\center
\includegraphics[width=\linewidth]{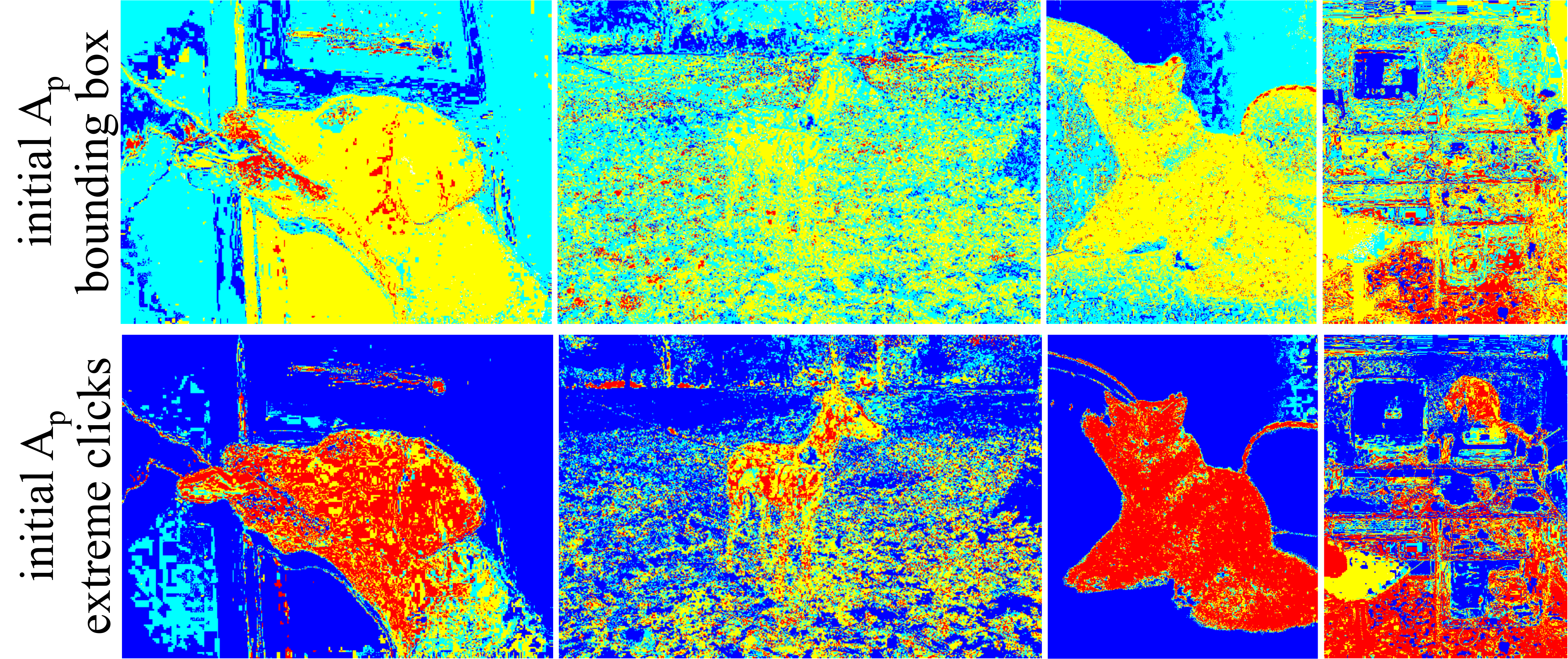}
\vspace{-.4cm}
\caption{\small Posterior probability of pixels belonging to object. For both rows the background appearance model is created by using an area outside the initial box (see Fig.~\ref{fig:boundApp}). In the first row the object model is created using the area inside the initial box. In the second row the object model is created from the object surface estimated using extreme clicks (Fig.~\ref{fig:boundApp}, third row in light-green). Predictions from the appearance model using extreme clicks are visibly better.}
\vspace{-.4cm}
\label{fig:appModels}
\end{figure}

However, because we have extreme clicks we can do better. We use them to obtain an initial object surface estimate (described above) from which we initialize the object appearance model. Fig.~\ref{fig:appModels} illustrates how this improves the unary potentials $U$ resulting from the appearance models.

\mypar{Clamping pixels.}

GrabCut sometimes decides to label all pixels either as object or background. To prevent this, one can clamp some pixels to a certain label. For the background, all pixels outside the bounding box are typically clamped to background.
For the object, one possible approach is to clamp a small area in the center of the box~\cite{ferrari08cvpr}.
However, there is no guarantee that the center of the box is on the object, as many objects are not convex.
Moreover, the size of the area to be clamped is not easy to set.


In this paper, we estimate the pixels to be clamped by skeletonizing the object surface estimate derived from our extreme clicks (described above).
In Sec.~\ref{sec:resultsSeg} we show how our proposed object appearance model initialization and clamping scheme affect the final segmentation quality.

\mypar{Pairwise potential $V$.}

The summation over $(p, q)$ in \eqref{eq:energyGC} is defined on an eight-connected pixel grid. Usually, this penalty depends on the RGB difference between pixels, being smaller in regions of high contrast~\cite{BoykovICCV01,BlakeECCV04,gulshan10cvpr,lempitsky:iccv09,Rother04vitto,veksler08eccv}. In this paper, we instead use the sum of the edge responses of the two pixels given by the edge detector~\cite{dollar13iccv}. In Sec.~\ref{sec:resultsSeg} we evaluate both pairwise potentials and show how they affect the final segmentation.

\mypar{Optimization.}

After the initial estimation of appearance models, we follow~\cite{Rother04vitto} and alternate between finding the optimal segmentation $L$ given the appearance models, and updating the appearance models given the segmentation.
The first step is solved globally optimally by minimizing~\eqref{eq:energyGC} using graph-cuts~\cite{boykov:pami04}, as our pairwise potentials are submodular. The second step simply fits GMMs to labeled pixels.

%% file: secResultsTradeoff.tex
\section{Extreme Clicking Results}
\label{sec:resultsDet}


\begin{table*}[t]
\vspace{-.3cm}
\centering
\footnotesize
\begin{tabular}{|c|c||c|c|c||c|c||c|c|} \hline
 &  & \multicolumn{3}{|c||}{Annotation quality w.r.t. GT SegBoxes} & \multicolumn{2}{|c||}{Detector performance (mAP)}
 & \multicolumn{2}{|c|}{Annotation time} \\ 
 \hline
Dataset & Annotation approach & mIoU & IoU$>$0.7 & IoU$>$0.5 & AlexNet & VGG16 & dataset (h) &
instance (s) \\
\hline
PASCAL & Extreme clicks & 88 & 92 & 98 & 56 & 66 & 14.3 & 7.0 \\
VOC 2007 & PASCAL GT Boxes & 88 & 93 & 98 & 56 & 66 & 70.0 & 34.5 \\
\hline \hline
PASCAL & Extreme clicks & 87 & 91 & 95 & 52 & 62 & 16.8 & 7.2 \\
VOC 2012 & PASCAL GT Boxes & 87 & 90 & 96 & 52 & 62 & 79.8 & 34.5 \\
\hline
\end{tabular}
\caption{\small Comparison of extreme clicking and PASCAL VOC ground-truth.}
\vspace{-.2cm}
\label{tab:objDetHumanAgreement}
\end{table*}

\begin{table*}[t]
\centering
\footnotesize
\begin{tabular}{|c|c||c|c|c||c|c||c|c|} \hline
 &  & \multicolumn{3}{|c||}{Annotation quality w.r.t. GT Boxes} & \multicolumn{2}{|c||}{Detector performance (mAP)}
 & \multicolumn{2}{|c|}{Annotation time} \\ 
 \hline
Dataset & Annotation approach & mIoU & IoU$>$0.7 & IoU$>$0.5 & AlexNet & VGG16 & dataset (h) &
instance (s) \\
\hline
 & Extreme clicks & 88 & 94 & 97 & 56 & 66 & 14.3 & 7.0 \\
PASCAL VOC & Human verification~\cite{papadopoulos16cvpr} & -- & -- & 81 & 50 & 58 & 9.2 & 4.5 \\
2007 & WSOL: Bilen and Vedaldi~\cite{bilen16cvpr} & -- & -- & 54 & 35 & 35 & 0 & 0 \\
\hline \hline
ILSVRC (subset) & box drawing in~\cite{russakovsky15cvpr} & -- & 71 & -- & -- & -- & -- & 12.3 \\
\hline
\end{tabular}
\caption{\small Comparison of extreme clicking and alternative fast annotation approaches.}
\vspace{-.5cm}
\label{tab:objDetMAP}
\end{table*}

We implement our annotation scheme on Amazon Mechanical Turk (AMT) and collect extreme click
annotations for both the trainval set of PASCAL VOC 2007~\cite{pascal07:thomas} (5011 images) and the
training set of PASCAL VOC 2012~\cite{pascal-voc-2012} (5717 images), which contain 20 object
categories. For every image we annotate a single instance per class (if present in the image), which enables direct comparison to other methods described below.
We compare methods both in terms of efficiency and quality.

\mypar{Compared methods.}

Our main comparisons are to the existing ground-truth bounding boxes of PASCAL VOC. As discussed in Sec.~\ref{sec:relwork}, we use 34.5s as the reference time necessary to produce one such high quality bounding box by drawing it the traditional way~\cite{su12aaai}.


At the other extreme, it is possible to obtain lower quality bounding boxes automatically at zero
extra costs by using weakly supervised methods, which only input image-level labels.
We compare to the recent method of~\cite{bilen16cvpr}.

We also compare to two methods which strike a trade-off between accuracy and efficiency~\cite{russakovsky15cvpr,papadopoulos16cvpr}. In~\cite{russakovsky15cvpr}, manual box drawing is part of a complex computer-assisted annotation system. Papadopoulos et al.~\cite{papadopoulos16cvpr} propose an annotation scheme that only requires annotators to verify boxes automatically generated by a learning algorithm.
Importantly, both~\cite{papadopoulos16cvpr,russakovsky15cvpr} report both annotation time and quality, enabling proper comparisons.

\mypar{Evaluation measures.}
For evaluating efficiency we report time measurements, both in terms of annotating the whole dataset
and per instance.

We evaluate the quality of bounding boxes with respect to the PASCAL VOC ground-truth.
We do this with respect to the ground-truth bounding boxes (\emph{GT Boxes}), but also with respect to
bounding boxes which we fit to the ground-truth
\emph{segmentations} (\emph{GT SegBoxes}).
We quantify quality by intersection-over-union (IoU)~\cite{everingham10ijcv}, where we measure the percentage of bounding boxes we annotated per object class with IoU greater than 0.5 and 0.7, and then take the mean over all classes (IoU$>$0.5, IoU$>$0.7). In addition, we calculate the average IoU for all instances of a class and take the
mean over all classes (mIoU).

As an additional measure of accuracy we measure detector performance using Fast-RCNN~\cite{Girshick15iccv}, trained either on our extreme click boxes or on the PASCAL GT Boxes.

\subsection{Results on quality and efficiency}


\vspace{0.5cm}
\mypar{PASCAL ground-truth boxes vs. extreme clicks.}
Table~\ref{tab:objDetHumanAgreement} reports the results.
Having two sets of ground-truth boxes enables us to measure the agreement among the expert annotators that created PASCAL. Comparing GT Boxes and GT SegBoxes reveals this agreement to be at 88\% mIoU on VOC 2007. Moreover, 93\% of all GT Boxes have IoU $>0.7$ with their corresponding GT SegBox. This shows that the ground-truth annotations are highly consistent, and these metrics represent the quality of the ground-truth itself. Similar findings apply to VOC 2012.

Interestingly, the boxes derived from our extreme clicks achieve equally high metrics, when compared to the PASCAL ground-truth annotations. Therefore our extreme click annotations yield boxes with a quality within the agreement among
expert-annotators using the traditional way of drawing. To get a better feeling for such quality, if we perturb each of
the four coordinates of the GT Boxes by 4 pixels, the resulting boxes also have 88\% mIoU with the
unperturbed annotations. Qualitative examples are shown in the appendix~\ref{sec:appendix}. 

To further demonstrate the quality of extreme clicking, we train Fast-RCNN~\cite{Girshick15iccv} using either PASCAL GT Boxes or extreme click boxes. We train on PASCAL VOC 2007’s trainval set and test on its test set, then we train on VOC 2012’s train and test on its val set. We experiment using AlexNet~\cite{krizhevsky12nips} and VGG16~\cite{simonyan15iclr}.
Performance when training from GT Boxes or from our boxes is identical on both datasets and using both base networks.

\mypar{Annotation efficiency.}
In terms of annotation efficiency, extreme clicks are $5\times$ cheaper: 7.0s instead of 34.5s.
This demonstrates that extreme clicking costs only a fraction of the annotation time of the widely used box-drawing
protocol~\cite{crowdflowerboxannotation16,everingham10ijcv,russakovsky15cvpr,spare5boxannotation17,su12aaai},
without any compromise on quality.

\mypar{Human verification~\cite{papadopoulos16cvpr} vs. extreme clicks.}
Table~\ref{tab:objDetMAP} compares extreme clicks to human
verification~\cite{papadopoulos16cvpr} on VOC 2007. While verification is $1.6\times$ faster, our bounding
boxes are much more accurate (97\% correct at IoU$>$0.5, compared to 81\%
for~\cite{papadopoulos16cvpr}). Additionally, detector performance at test time is 6\%-8\% mAP 
higher for extreme clicking.

\mypar{Weak supervision vs. extreme clicks.}
Weakly supervised methods are extremely cheap in human supervision time. However, the recent work~\cite{bilen16cvpr} reports 35\% mAP using VGG16, which is only about half the result brought by extreme clicking (66\% mAP, Table~\ref{tab:objDetMAP}).

\mypar{Box drawing~\cite{russakovsky15cvpr} vs. extreme clicks.}
Finally, we compare to~\cite{russakovsky15cvpr} in Table~\ref{tab:objDetMAP}. This is an approximate comparison as
measurements of their box-drawing component are done on an unspecified subset of ILSVRC 2014.
However, as ILSVRC and PASCAL VOC are comparable in both quality of annotations and difficulty of
the dataset~\cite{russakovsky15ijcv}, this comparison is representative.
%
%
In~\cite{russakovsky15cvpr} they report 12.3s for drawing a bounding box, where 71\% of the drawn
boxes have an IoU$>$0.7 with the ground-truth box. This suggests that bounding boxes can be drawn faster than reported in~\cite{su12aaai} but this comes with a significant drop in quality.
In contrast, extreme clicking costs 7s per box and 91\%-94\% of those boxes have IoU$>$0.7. Hence our protocol to
annotate bounding boxes is both faster and more accurate.


\subsection{Additional analysis}



\vspace{3mm}
\mypar{Per-click response-time.}
We examine the mean response time per click during extreme clicking. Interestingly, the first click
on an object takes about 2.5s, while subsequent clicks take about 1.5s. This is because the annotator needs to find the object in the image before they can make the first
click. Interestingly, 1s visual search is consistent with earlier findings~\cite{ehinger17vico,papadopoulos14eccv}.

\mypar{Influence of qualification test and quality control.}
We conducted three crowd-sourcing experiments on 200 trainval images of PASCAL VOC 2007 to test the influence of using a qualification test and quality control. We report the quality of the bounding boxes derived from extreme clicks in Tab.~\ref{tab:mturkInflQQ}.
Using a qualification test vastly improves annotation quality (from 75.4\% to 85.7\% mIoU).
The quality control brings a smaller further improvement to 87.1\% mIoU.

\begin{table}[t]
\centering
\footnotesize
\begin{tabular}{|c|c||c|c|} \hline
Qualification test & Quality control &  mIoU & IoU$>$0.7\\
\hline
 &  &  75.4 & 68.0\\
\checkmark &  & 85.7 & 91.0 \\
\checkmark & \checkmark & 87.1 & 92.5 \\
\hline
\end{tabular}
\caption{\small Influence of the qualification test and quality control on the accuracy of extreme click annotations (on 200 images from PASCAL VOC 2007).}
\vspace{-.4cm}
\label{tab:mturkInflQQ}
\end{table}

\mypar{Actual Cost.} We paid the annotators $\$0.15$ to annotate a batch of 10 images which, based
on our timings, is about $\$7.7$ per hour. The total cost for annotating the whole trainval set of
PASCAL VOC 2007 and the training set of PASCAL VOC 2012 was $\$147$ and $\$167$, respectively.

%% file: secResultsSeg.tex
\section{Results on Object Segmentation}
\label{sec:resultsSeg}

This section demonstrates that one can improve segmentation from a bounding box by using also the boundary points which we obtain from extreme clicking.


\subsection{Results on PASCAL VOC}

\vspace{-.1cm}
\paragraph{Datasets and Evaluation.}
We perform experiments on VOC 2007 and VOC 2012. The trainval set of the segmentation
task of VOC 2007 consists of 422 images with ground-truth segmentation masks of 20 classes.
For VOC 2012, we evaluate on the training set, using as reference ground-truth the augmented masks set 
by~\cite{hariharan11iccv} (5623 images).

To evaluate the output object segmentations, for every class we compute the intersection over
union (IoU) between the predicted and ground-truth segmentation mask, and report the mean IoU over all object
classes (mIoU). Some pixels in VOC 2007 are labeled as `unknown' and are excluded from evalutation.
For these experiments we use structured edge forests~\cite{dollar13iccv} to predict object
boundaries, which is trained on BSD500~\cite{arbelaez11pami}.


\mypar{GrabCut from PASCAL VOC GT Boxes.}
We start with establishing our baseline by using GrabCut on the original GT Boxes of VOC (for
which no boundary points are available). Since applying~\cite{Rother04vitto} directly leads to rather
poor performance on VOC 2007 (37.3\% mIoU), we first optimize GrabCut on this dataset using methods discussed in
Sec.~\ref{sec:segMethod}. Our optimized model has the following properties: the object appearance
model is initialized from all pixels within the box. The background appearance model is initialized
from a small ring around the box which has twice the area of the bounding box.
A small rectangular core centered within the box whose area is a quarter of the area of the box is clamped to be object. All pixels outside the box are clamped to be background. As pairwise
potential, instead of using standard RGB differences, we use the summed edge responses of
~\cite{dollar13iccv} of the corresponding pixels. All modifications together substantially improve results to 74.4\% mIoU on VOC 2007. We then run GrabCut again on VOC 2012 using the exact same settings optimized for VOC 2007, obtaining 71.0\% mIoU.




\mypar{GrabCut from extreme clicking.}
Thanks to our extreme clicking annotations, we also have object boundary points. Starting from the optimized
GrabCut settings established in the previous paragraph, we make use of these boundary points to (1)
initialize a better object appearance model, and (2) choose better pixels to clamp to object.
As described in Sec.~\ref{sec:segMethod}, we use the extreme clicks to estimate an initial
contour of the object by following predicted object boundaries~\cite{dollar13iccv}. We use the surface
bounded by this contour estimate to initialize the appearance model. We also skeletonize this
surface and clamp the resulting pixels to be object.
The resulting model yields 78.1\% mIoU on VOC 2007 and 72.7\% on VOC 2012. This is an improvement of 3.7\% (VOC 2007) and 1.7\% (VOC 2012) over the strong baseline we built.
Fig.~\ref{fig:boundApp} shows qualitative results comparing GrabCut segmentations starting from GT Boxes (last row) and those based on our extreme clicking annotations (second-last row).

\subsection{Results on the GrabCut dataset}

We also conducted an experiment on the Grabcut dataset~\cite{Rother04vitto}, consisting of only 50
images. The standard evaluation measure is the error rate in terms of the percentage of mislabelled
pixels. For this experiment, we simulate the extreme click annotation by using the extreme points of
the ground-truth segmentation masks of the images. 

When we perform GrabCut from bounding boxes, we obtain an error rate of 8\%.
When using additionally the boundary points from simulated extreme clicking, we obtain
5.5\% error, an improvement of 2.5\%. This again demonstrates that boundary points contain useful
information over bounding boxes alone for this task.

For completeness, we note that the state-of-the-art method on this dataset has 3.6\% error~\cite{wu14cvpr}. This method
uses a framework of superpixels and Multiple Instance Learning to turn a bounding box into a
segmentation mask.
In this paper we build on a much simpler segmentation framework (GrabCut).
We believe that incorporating our extreme clicks into~\cite{wu14cvpr} would bring further improvements.

\subsection{Training a semantic segmentation model}

We now explore training a modern deep learning system for semantic segmentation from the segmentations derived from extreme clicking.  We train DeepLab~\cite{chen15iclr,papandreou15iccv} based on VGG-16~\cite{simonyan15iclr} on the VOC
  2012 train set (5,623 images) and then we test on its val set (1,449 images). We
  measure performance using the standard mIoU measure (Tab.~\ref{tab:segTrain}). We compare our approach to full supervision 
  by training on the same images but using the ground-truth, manually drawn object segmentations (one instance per class per image, for fair comparison). We also compare to training on segmentations generated from GT Boxes.
\mypar{Full supervision} yields 59.9\% mIoU, which is our upper bound. As a reference, training on manual segmentations for all instances in the dataset yields 63.8\% mIoU. This is 3.8\% lower than in~\cite{papandreou15iccv} since they train from train+val using the extra annotations by~\cite{hariharan11iccv} (10.3k images).
\mypar{Segments from GT Boxes} result in 55.8\% mIoU.
\mypar{Segments from extreme clicks} lead to 58.4\% mIoU. This means our extreme clicking segmentations lead to a +2.6\% mIoU improvement over those generated from bounding boxes. Moreover, our result is only -1.5\% mIoU below the fully supervised case (given the same total number of training samples).

 \begin{table}[t]
 \centering
 \footnotesize
 \begin{tabular}{|c|c|c|c|} \hline
 & Full & Segments from & Segments from\\
 & supervision & GT Boxes & extreme clicks \\
 \hline
 mIoU & 59.9 & 55.8 & 58.4 \\
 \hline
 \end{tabular}
 \caption{\small Segmentation performance on the val set of PASCAL VOC 2012 dataset using different types of annotations.} 
 \vspace{-.4cm}
 \label{tab:segTrain}
 \end{table}

%
%

%% file: secConclusions.tex
\section{Conclusions}
\label{sec:conclusions}

We presented an alternative to the common way of drawing
bounding boxes, which involves clicking on imaginary corners of an imaginary box. Our alternative is
extreme clicking: we ask annotators to click on the top, bottom, left- and right-most points of an
object, which are well-defined physical points. 
We demonstrate that our method delivers bounding boxes that are as good as traditional drawing,
while taking just 7s per annotation. To achieve this same level of
quality, traditional drawing needs 34.5s~\cite{su12aaai}. Hence our method cuts annotation costs by a factor $5\times$ without any compromise on quality.

In addition, extreme clicking leads to more than just a box: we also obtain accurate object boundary points. To demonstrate their usefulness we incorporate them into GrabCut, and show
that they leads to better object segmentations than when initializing it from the bounding box alone.
Finally, we have shown that semantic segmentation models trained on these segmentations perform close to those trained with manually drawn segmentations (when given the same total number of samples). 

\mypar{Acknowledgement.}
 This work was supported by the ERC Starting Grant ``VisCul''.

%% file: secAppendix.tex
\begin{figure*}[t]
\center
\includegraphics[width=\linewidth]{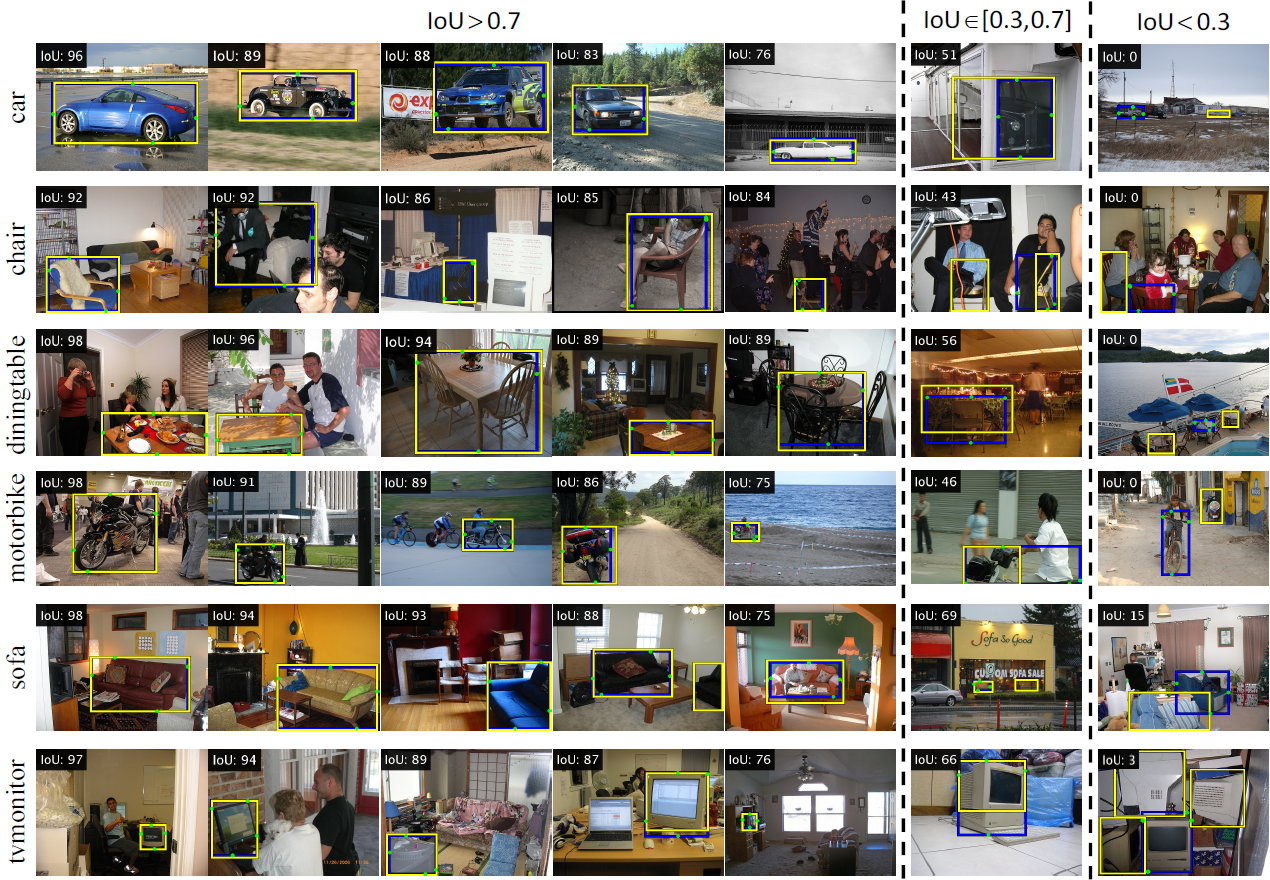}
\vspace{-.4cm}
\caption{\small \textbf{Qualitative examples of extreme clicking annotations on PASCAL VOC 2007 trainval set.} For each example, we provide the extreme clicking bounding-box (blue box) and the exact positions of the extreme clicks (green dots), the PASCAL ground-truth bounding-boxes (yellow box) and the exact IoU between the two annotations.}
\label{fig:qualClick}
\end{figure*}

\section*{Appendix}
\appendix
\section{Qualitative examples of extreme clicking}
\label{sec:appendix}

Fig.~\ref{fig:qualClick} shows qualitative results on various PASCAL VOC
2007 trainval images comparing extreme clicking with the original ground-truth (GT Boxes). As shown in Tab.~\ref{tab:objDetMAP}, 94\% of the extreme clicking boxes
have high IoU (IoU$>$0.7) with the corresponding GT Box. Only for 6\% of all objects the two annotations are considerably different (IoU$<$0.7).
In order to fully understand why in these few cases the extreme clicks and PASCAL annotations diverge, we manually inspected annotations for 50 randomly picked samples with almost no overlap (IoU$<$0.3) and 100 samples with low overlap (IoU $\in$ [0.3,0.7]).

We found out that in cases with almost no overlap, 62\% of our annotations are correct but those objects are not annotated in the PASCAL ground-truth; 18\% annotations are on objects of a similar class (e.g. a side-table instead of a dining-table, a vase with flowers instead of a potted-plant, a pickup truck instead of a car); 14\% are on an entirely wrong class; 6\% are spatial annotation errors (e.g. missing a part of the object).

For low overlap cases there are errors either in extreme click annotations or in the PASCAL annotations. The majority of these cases are partially occluded objects, where small parts of an object (e.g. hand, foot, tail, leg, wheel) appear quite far from the main visible part. Other cases are objects with thin parts like antennas and masts, yet others are dark images. In 21\% of these low overlap cases, extreme clicks provided better annotations, in 23\% the PASCAL ground-truth was better, and in 56\% of the cases we could not decide which annotation was better (these were mostly small objects).